\documentclass[pdflatex,sn-mathphys-num]{sn-jnl}
\usepackage{}
\usepackage{multirow}
\usepackage{amsmath,amssymb,amsfonts}
\usepackage{amsthm}
\usepackage{mathrsfs}
\usepackage[title]{appendix}
\usepackage{xcolor}
\usepackage{float}
\usepackage{textcomp}
\usepackage{manyfoot}
\usepackage{booktabs}
\usepackage{algorithm}
\usepackage{algorithmicx}
\usepackage{algpseudocode}
\usepackage{listings}
\usepackage{stfloats}
\usepackage{url}
\usepackage{verbatim}
\usepackage{natbib}
\usepackage{siunitx}
\usepackage[utf8]{inputenc}
\usepackage{adjustbox}
\usepackage[labelfont=bf]{caption} 
\usepackage{subcaption}   

\usepackage{threeparttable}
\usepackage{placeins}
\usepackage{float}
\usepackage{tipa}
\theoremstyle{thmstyleone}%
\theoremstyle{thmstyletwo}%
\theoremstyle{thmstylethree}%

\usepackage[table]{xcolor}
\definecolor{tablebeige}{RGB}{250,247,238}
\expandafter\ifx\csname pdfobjcompresslevel\endcsname\relax\else
\fi
\newcounter{suppfigcounter}
\usepackage{amsthm}
\usepackage{mathrsfs}

\usepackage{multirow}
\usepackage{siunitx}
\usepackage{adjustbox}
\usepackage{threeparttable}
\usepackage{float}
\usepackage{placeins}
\usepackage{cuted}

\usepackage[labelfont=bf]{caption}

\usepackage{textcomp}
\usepackage{manyfoot}
\usepackage{algorithm}
\usepackage{algorithmicx}
\usepackage{algpseudocode}
\usepackage{listings}
\usepackage[title]{appendix}
\usepackage{verbatim}
\usepackage{tipa}
\usepackage{url}
\usepackage{natbib}
\usepackage{hyperref}
\theoremstyle{thmstyleone}
\theoremstyle{thmstyletwo}
\theoremstyle{thmstylethree}
\newcommand{\suppfigsection}[2]{%
    \refstepcounter{suppfigcounter}
    \subsection*{ Extended Data Figure \thesuppfigcounter: #1}%
    \label{#2}%
}
\expandafter\ifx\csname pdfobjcompresslevel\endcsname\relax\else
\fi

\begin{document}

\title[Algorithm-hardware co-design of neuromorphic networks with dual memory pathways]{Algorithm-hardware co-design of neuromorphic networks with dual memory pathways}
\author[1]{\fnm{Pengfei} \sur{Sun}}\email{p.sun@imperial.ac.uk}
\equalcont{These authors contributed equally to this work.}
\author[2]{\fnm{Zhe} \sur{Su}}\email{zhesu@ini.ethz.ch}
\equalcont{These authors contributed equally to this work.}
\author[3]{\fnm{Jascha} \sur{Achterberg}}  \email{jascha.achterberg@dpag.ox.ac.uk}
\author[2]{\fnm{Giacomo} \sur{Indiveri}}
\email{giacomo@ethz.ch}
\author[1]{\fnm{Dan F.M.} \sur{Goodman}}
\email{d.goodman@imperial.ac.uk}
\author[1,4,5]{\fnm{Danyal} \sur{Akarca}}\email{d.akarca@imperial.ac.uk}

\affil[1]{\orgdiv{Department of Electrical and Electronic Engineering}, \orgname{Imperial College London}}
\affil[2]{\orgdiv{Institute of Neuroinformatics}, \orgname{University of Zurich and ETH Zurich}}
\affil[3]{\orgdiv{Centre for Neural Circuits and Behaviour}, \orgname{University of Oxford}}
\affil[4]{\orgdiv{Imperial-X}, \orgname{Imperial College London}}
\affil[5]{\orgdiv{MRC Cognition and Brain Sciences Unit}, \orgname{University of Cambridge}}


\abstract{

  Spiking neural networks excel at event-driven sensing. Yet, maintaining task-relevant context over long timescales both algorithmically and in hardware, while respecting both tight energy and memory budgets, remains a core challenge in the field.
  We address this challenge through an algorithm-hardware co-design effort.
  At the algorithm level, inspired by the cortical fast-slow organization in the brain, we introduce a neural network with an explicit slow memory pathway that, combined with fast spiking activity, enables a dual memory pathway (DMP) architecture in which each layer maintains a compact low-dimensional state that summarizes recent activity and modulates spiking dynamics.
  This explicit memory stabilizes learning while preserving event-driven sparsity, achieving competitive accuracy on long-sequence benchmarks with 40-60\% fewer parameters than equivalent state-of-the-art spiking neural networks.
  At the hardware level, we introduce a near-memory-compute architecture that fully leverages the advantages of the DMP architecture by retaining its compact shared state while optimizing dataflow, across heterogeneous sparse-spike and dense-memory pathways.
  We show experimental results that demonstrate more than a 4$\times$ increase in throughput and over a 5$\times$ improvement in energy efficiency compared with state-of-the-art implementations.
  Together, these contributions demonstrate that biological principles can guide functional abstractions that are both algorithmically effective and hardware-efficient, establishing a scalable co-design framework for real-time neuromorphic computation and learning.
}

\maketitle

\section{Introduction}

Brains solve temporal problems in real time and under tight energy budgets, that are unmatched in artificial systems. One hypothesis for the brain's superior temporal processing abilities is that their computation is both continuous time and event-driven, based on discrete sparse spiking. Spiking neural networks (SNNs) inherit these two core principles, making them a promising and biologically grounded complement to dense artificial neural networks (ANNs) \citep{maass1997networks,roy2019towards}. Yet one brain-like ability critically important for temporal processing remains hard to reproduce efficiently: retaining information over behaviorally relevant timescales. Standard leaky integrate-and-fire (LIF) neurons integrate spikes into a decaying membrane potential \citep{GerstnerKistler2002} which captures instantaneous evidence but progressively suppresses older inputs over time scales of milliseconds \citep{li2024brain,yu2025beyond}. Conversely, precise temporal codes (e.g., time-to-first-spike) preserve timing but do not carry graded intensity over long windows \citep{mostafa2017supervised,comcsa2021temporal,sun2024delay}. Another way of thinking about this is that a purely feedforward SNN typically captures either \emph{what} happened or \emph{when} it happened, but not both. The fundamental challenge is therefore maintaining long-range temporal context without sacrificing event-driven efficiency.

 \begin{figure}
     \centering
     \includegraphics[width=\linewidth]{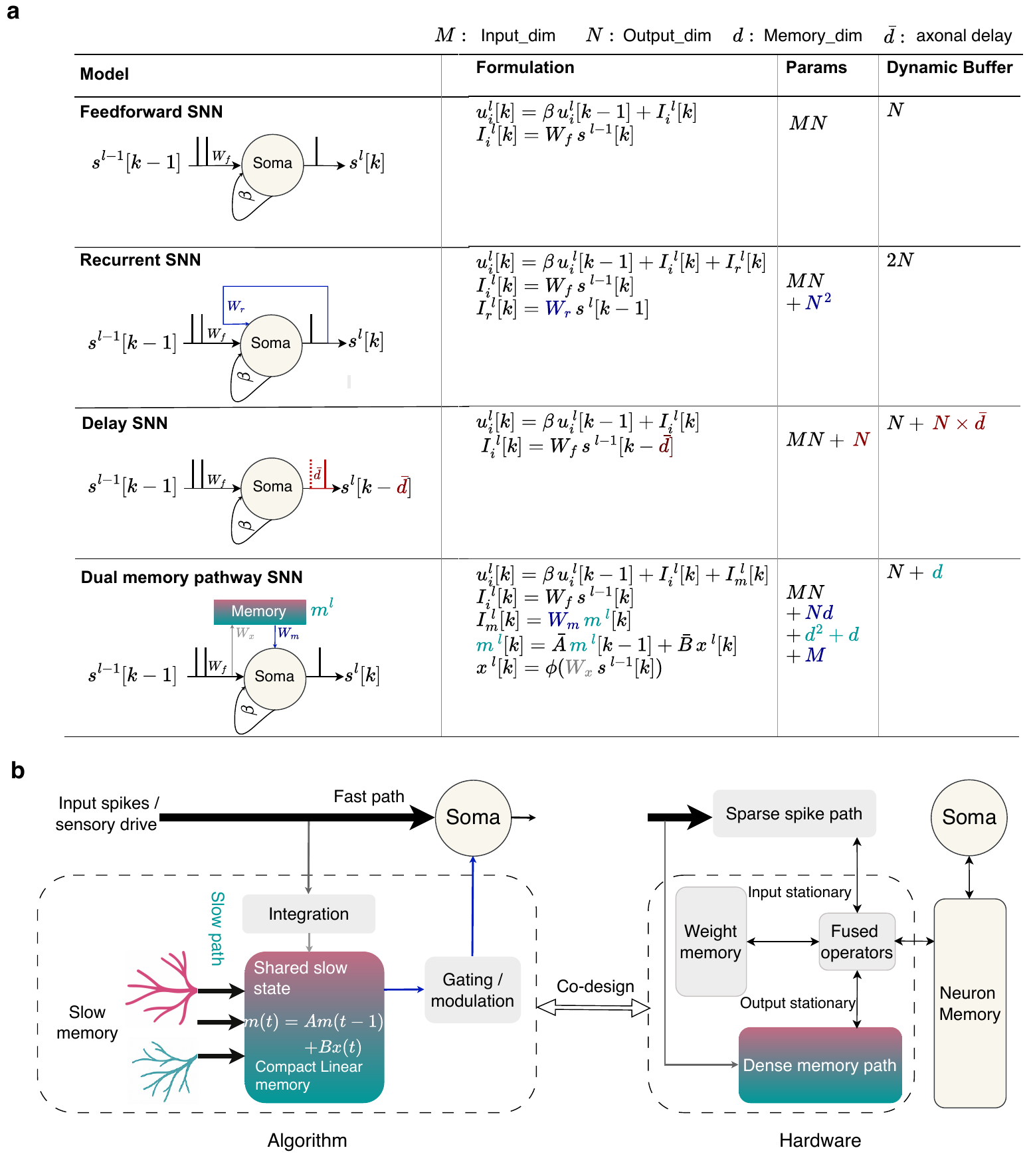}
\caption{\textbf{From fast-slow cortical motifs to the dual memory pathway architecture.}
\textbf{a} Schematic comparison of Feedforward SNN, Recurrent SNN, Delay SNN, and the proposed Dual memory pathway SNN, highlighting their discrete-time formulations, parameter counts, and dynamic buffer requirements. \(M,N\) denote input/output widths and \(d\) the memory dimension (\(d\!\ll\!N\)). \textbf{b} Dual memory pathway abstraction: at the algorithmic level (left), each layer maintains a shared, low-dimensional state that captures slow contextual dynamics and modulates fast spiking activity; at the hardware level (right), this separation is mirrored by a heterogeneous accelerator that keeps the compact state on-chip and fuses sparse and dense computations for efficient execution.}

  \label{fig:first}
\end{figure}

Several strategies have been explored to extend temporal capacity in SNNs \citep{gast2024neural}. One approach augments recurrence, often via heterogeneous time constants or adaptive thresholds \citep{fang2021incorporating,perez2021neural,shaban2021adaptive,yin2021accurate,bellec2018long}, allowing activity to persist through feedback. Another exploits biologically inspired transmission delays (axonal, synaptic, or dendritic) to align spikes to task-relevant timescales and act as implicit temporal buffers \citep{sreenivasan2019and,izhikevich2006polychronization, sun2023learnable,wang2019delay,yu2022improving,grappolini2023beyond,hammouamrilearning,10181778,d2024denram,zheng2024temporal,meszaros2025efficient}. Both strategies improve temporal integration, yet each imposes substantial implementation costs in hardware. Dense recurrence scales quadratically with layer width, erodes sparsity benefits, and requires continuous recirculation of activity which increases both the memory footprint and energy consumption. Long, learnable delays avoid explicit recurrence but can demand deep on-chip buffers and per-connection timing metadata, inflating area, power, and latency in neuromorphic implementations \citep{patino2024hardware,karilanova2025delays}. {We also note that reservoir-based spiking approaches, exemplified by liquid state machines, exploit rich transient dynamics in a fixed recurrent substrate and learn only a simple readout, offering an alternative route to temporal processing~\cite{maass2002realtime}. However, in more challenging real-world settings their accuracy can be limited, particularly under constrained hardware deployment~\cite{biswas2024temporal,deckers2022extended}.} This creates a design challenge: temporal memory mechanisms must be sufficiently expressive to capture long-range dependencies while remaining compatible with efficient neuromorphic implementation. {Addressing this requires reconsidering both algorithmic structure and hardware design \citep{thakur2018large,roy2019towards, lenz2020event,payvand2020chip, wu2022brain, dampfhoffer2023backpropagation, zheng2025modeling}.}

The mammalian cortex inspires a potential architectural solution. Cortical circuits maintain efficient long-range temporal context through dendritic branches that integrate inputs over a spectrum of timescales, while neuronal populations exhibit correspondingly heterogeneous temporal constants that shape fast somatic spiking without requiring dense global recurrence \citep{london2005dendritic,murray2014hierarchy, zheng2024temporal}. Recent evidence suggests that cortical networks dynamically recruit fast and slow computational pathways according to task demands, revealing a functional organization built around multiple interacting timescales \cite{sartzetaki2025human,cook2025brainlike}. We distill this into a functional abstraction: pair fast spiking dynamics with compact temporal memory. Following this abstraction, we introduce a memory-augmented spiking architecture in which each layer maintains a low-dimensional state vector \( m \in \mathbb{R}^d \), with \( d \ll N \) for a layer of \( N \) spiking neurons. {
This state evolves under well-conditioned slow dynamics and feeds back as an additional input current to the neurons, acting as a compact {linear} slow pathway that provides shared context and modulates fast spiking within each layer. This deployment-oriented fast/slow decomposition enables controllable trade-offs between memory footprint, and accuracy under tight resource budgets.} Rather than storing full spike histories or broadcasting dense recurrent activity, the network compresses recent activity into a few slow modes and exposes that compressed context at each timestep. {This dual memory pathway (DMP) architecture is motivated by long-horizon temporal memory under stringent hardware constraints, and is thus tailored to regimes in which long-range temporal dependencies are the primary limiting factor.
}

To fully exploit the advantages afforded by this memory architecture, we introduce a digital near-memory-compute architecture that employs heterogeneous dataflow optimization to maximize arithmetic intensity. This design overcomes limitations of prior hardware, in which conventional recurrent SNN accelerators must scale both parameter storage and computation quadratically \cite{reckon}, while hardware supporting long learnable delays requires deep buffers in digital implementations \cite{loihi2} or large capacitors in mixed-signal designs \cite{d2024denram}, both significantly increasing energy consumption and chip area. The algorithmic structure we propose inherently enables a low-memory-footprint design that avoids these costs. Through algorithm-hardware co-design, we empirically observe that our system matches or outperforms recurrent and delay-based SNNs on event-driven auditory classification and long-horizon sequential inference, all while using only a small number of memory states. Post-layout simulations of the proposed hardware architecture in an advanced 22FDX technology demonstrate more than $4\times$ higher throughput and over $5\times$ better energy efficiency compared with SOTA designs.

Overall, we show how efficient temporal processing can be achieved through hardware-algorithm co-design inspired by circuit motifs in the brain. Imposing hardware considerations at the algorithmic level results in a network architecture with a temporal memory that is explicit and low-rank. By designing hardware around explicit memory and applying carefully orchestrated optimizations, we obtain a heterogeneous dataflow structure that efficiently integrates fast and slow memory. Jointly, this algorithm and hardware pairing achieves benchmark-beating temporal processing capabilities alongside SOTA energetic efficiency. We open-source the hardware architecture to enable community adoption and further co-design exploration.

\section{Results}
\begin{figure}[!tbp]
\centering
\includegraphics[width=1.05\linewidth]{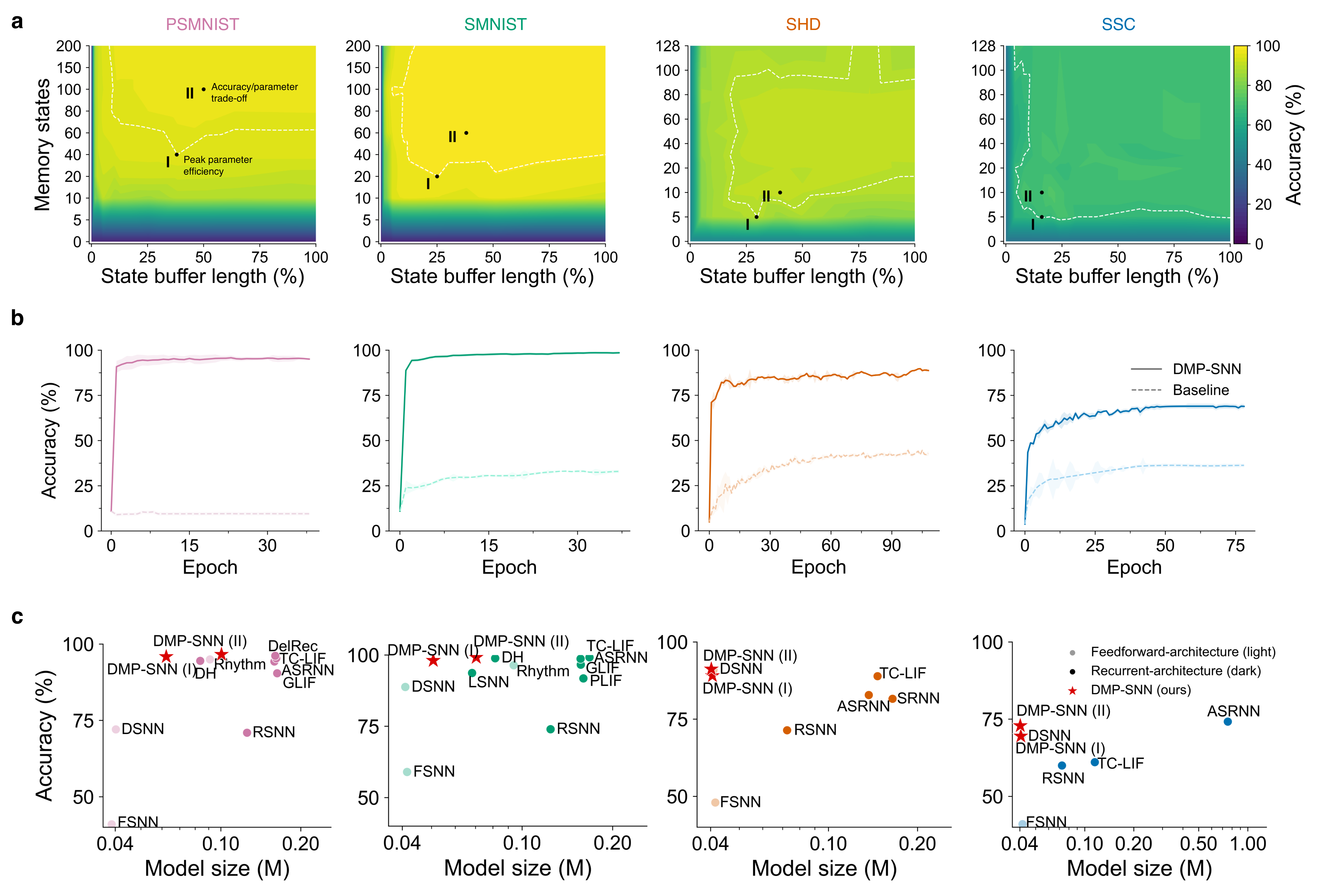}
\caption{\textbf{Accuracy-efficiency across temporally structured benchmarks.}
{\textbf{a} Test accuracy on PS-MNIST, SMNIST, SHD, and SSC for the proposed DMP-SNN. For all datasets, the {white dashed curve} denotes an {iso-accuracy  contour}, where each point corresponds to a (memory size, state-buffer length) pair achieving the same target accuracy (chosen to be competitive with prior work). The iso-accuracy contours indicate overall trends. The curve highlights two configurations of DMP-SNN: Solution~I, a {single operating point} on the curve-the most parameter-efficient configuration that matches state-of-the-art performance, and Solution~II (accuracy-parameter trade-off), a higher-capacity configuration with low additional parameter cost.}
\textbf{b} Learning curves for baseline and the proposed DMP-SNN on PS-MNIST, S-MNIST, SHD, and SSC. The central line indicates the mean accuracy across $n = 5$ independent runs, and the shaded area represents the standard error of the mean. Under identical training settings and neuron counts, the DMP-SNN starts from a higher accuracy and converges rapidly, consistently outperforming baseline across datasets.
\textbf{c} Comparison of DMP-SNN with recent strong SNN models. DMP-SNN matches or outperforms prior work while using a purely feedforward architecture and notably fewer parameters. We compared the recent FSNN\citep{cramer2020heidelberg}, DSNN\citep{sun2023learnable}, RSNN\citep{cramer2020heidelberg}, LSNN\citep{bellec2018long} , GLIF \citep{yao2022glif}, PLIF\citep{fang2021incorporating}, ASRNN \citep{yin2021accurate}, SRNN\citep{pagkalos2023introducing}, DH\citep{zheng2024temporal}, TC-LIF\citep{zhang2024tc}, Rhythm\citep{yan2025efficient}, and DelRec \citep{queant2025delrec}.
}

\label{fig:2}
\end{figure}

\subsection{Dual memory pathways for efficient temporal computation}

Feedforward spiking networks (FSNNs) are computationally efficient but perform poorly on tasks requiring long temporal dependencies, as their leaky membranes encode only transient evidence and rapidly forget past inputs. Recurrent SNNs (RSNN) extend temporal integration by feeding activity back within each layer, enabling longer retention but at the cost of quadratic scaling and high memory traffic that erode event-driven efficiency. Recent delay-based SNNs (DSNN), particularly those with learnable axonal delays, address this limitation by distributing memory across time, achieving remarkable performance on neuromorphic benchmarks with only linear parameter complexity (\(\mathcal{O}(N)\)) \cite{sun2025exploitingheterogeneousdelaysefficient}. However, these models rely on deep programmable buffers, which can increase hardware cost and latency.
Here, we introduce the dual memory pathway (DMP) architecture for spiking neural networks (DMP-SNN), which replaces these costly mechanisms with a compact, shared slow state that summarizes recent activity and feeds back as an additional input current (see \textbf{Figure \ref{fig:first}a}). Acting as an explicit low-dimensional working memory, at only 5\% of the hidden width, this state preserves the temporal context relevant to the task over behaviorally meaningful timescales while remaining fully compatible with efficient hardware implementation (\textbf{Figure \ref{fig:first}b}).

We evaluated this dual memory pathway architecture on two classes of sequential benchmarks. The first class targets long-range visual integration. In Sequential MNIST (S-MNIST), each $28\times28$ image is presented as a 784-step pixel stream, requiring integration across hundreds of steps. In Permuted Sequential MNIST (PS-MNIST), a fixed random permutation destroys local spatial structure, so performance depends almost entirely on long-range temporal memory \citep{lecun2002gradient}. The second class targets event-based auditory classification \citep{cramer2020heidelberg}. The Spiking Heidelberg Digits (SHD) dataset encodes spoken digits as 700-channel spike trains generated by a cochlear model, and the Spiking Speech Commands (SSC) dataset similarly represents 35 spoken-word categories. Together, these benchmarks span dense, clocked sequences and irregular, event-driven input, and probe temporal dependencies on different timescales.

We benchmarked four spiking network architectures: FSNN, RSNN, DSNN, and our DMP-SNN (\textbf{Table \ref{tbl:1}}). FSNNs fail on long-range structure (11.3\% on PS-MNIST). Recurrence in RSNNs improves temporal integration (71.0\% on PS-MNIST; 71.4\% on SHD) but scales as $\mathcal{O}(N^2)$. DSNNs perform well on auditory tasks (91\% on SHD; 70\% on SSC) by learning axonal timing, but require long programmable delays.  The DMP-SNN reaches 99.3\% on S-MNIST and 97.3\% on PS-MNIST. Our method matches delay-based models on SHD and SSC, and in the single-layer SSC setting outperforms them. \textbf{Figure \ref{fig:2}a} shows that even a small number of memory states yields consistent gains, and that once the state buffer length exceeds roughly 25\% of the sequence, performance is already competitive, indicating that the relevant temporal information has been captured. \textbf{Figure \ref{fig:2}b} illustrates the learning dynamics of the baseline model relative to the DMP-SNN, showing that memory-augmentation leads to both quicker convergence and higher accuracy. Crucially, this is achieved using only a low-dimensional auxiliary state with $\mathcal{O}(d)$ overhead, where $d \ll N$ (typically 5-10\% of the hidden width) can match or exceed SOTA performance with fewer parameters while providing long-timescale context (see solution I, the most parameter efficient solution, and solution II, the best performing solution, in \textbf{Figure \ref{fig:2}c}) \cite{bellec2018long, yin2021accurate, yao2022glif, zheng2024temporal,sun2024delayed, fan2025multisynaptic,  yan2025efficient}. { \textbf{Extended Data Table}~\ref{dvsgesture} further reports results on the DVS Gesture spatiotemporal event-stream benchmark, where incorporating DMP yields a consistent accuracy improvement with negligible parameter overhead.}

\begin{table}[!htbp]
\centering

\caption{\textbf{Performance on temporally structured benchmarks.} Test accuracy (\%) on PSMNIST, SMNIST, SHD, and SSC for four model families: FSNN, RSNN, DSNN, and the proposed DMP-SNN. For SSC, results are shown for networks with one and two hidden layers (1-layer / 2-layer). Beige rows mark DMP-SNN in three settings: Solution I (the most parameter-efficient, matching strong baselines), Solution II (accuracy-efficiency trade-off, achieving high accuracy and maintaining low parameter cost), and Peak accuracy. Chance-level accuracy is shown for reference.}

\label{tbl:1}

\begin{tabular}{l r c}
\toprule
\rowcolor{gray!20}
\textbf{Model}
& \textbf{Parameters}
& \textbf{Accuracy (\%)} \\
\midrule

\rowcolor{gray!10}
\multicolumn{3}{l}{\textbf{PS-MNIST}} \\

FSNN
& 42K
& 11.30  \\

RSNN
& 122K
& 71.00 \\

DSNN
& 43K
& 72.06 \\

DMP-SNN (\textbf{I})
&  61K
& \textbf{95.50} \\

DMP-SNN (\textbf{II})
& 102K 
& \textbf{96.65} \\

DMP-SNN (\textbf{Peak accuracy})
& 202K
& \textbf{97.32} \\

\addlinespace[6pt]

\rowcolor{gray!10}
\multicolumn{3}{l}{\textbf{S-MNIST}} \\

FSNN
& 42K
& 59.00 \\

RSNN
& 122K
& 74.00 \\

DSNN
& 43K
& 88.79 \\

DMP-SNN (\textbf{I})
& 51K
& \textbf{98.08} \\

DMP-SNN (\textbf{II})
& 73K
& \textbf{99.20} \\

DMP-SNN (\textbf{Peak accuracy})
& 202K
& \textbf{99.28} \\

\addlinespace[6pt]

\rowcolor{gray!10}
\multicolumn{3}{l}{\textbf{SHD}} \\

FSNN
& 37K
& 48.60 \\

RSNN
& 70K
& 71.40 \\

DSNN
& 37K
& 90.98 \\

DMP-SNN (\textbf{I})
&38K  &  \textbf{89.00} \\

DMP-SNN (\textbf{II})
& 40K
& \textbf{91.23} \\

DMP-SNN (\textbf{Peak accuracy})
& 46K
& \textbf{91.69} \\

\addlinespace[6pt]

\rowcolor{gray!10}
\multicolumn{3}{l}{\textbf{SSC}} \\

FSNN
& 22K/39K
& 26.08\,/\,38.50 \\

RSNN
& 39K/72K
& 50.90\,/\,60.00 \\

DSNN
& 23K/39K
& 60.01\,/\,69.40 \\

DMP-SNN (\textbf{I})
& 23K/40K
& \textbf{65.09\,/\,69.50} \\

DMP-SNN (\textbf{II})
& 24K/42K
& \textbf{65.37\,/\,72.90} \\

DMP-SNN (\textbf{Peak accuracy})
& 24K/42K
& \textbf{65.37\,/\,72.90} \\

\midrule
\multicolumn{3}{l}{Chance level:
10.00 (S-MNIST / PS-MNIST),
5.00 (SHD),
2.90 (SSC).} \\
\bottomrule
\end{tabular}
\begin{tablenotes}
\footnotesize
\item Bold font indicates this work.  
\end{tablenotes}

\end{table}

\subsection{Context-dependent temporal demands reveal design trade-offs}

\begin{figure}[!tbp]
\centering
\includegraphics[width=1.05\linewidth]{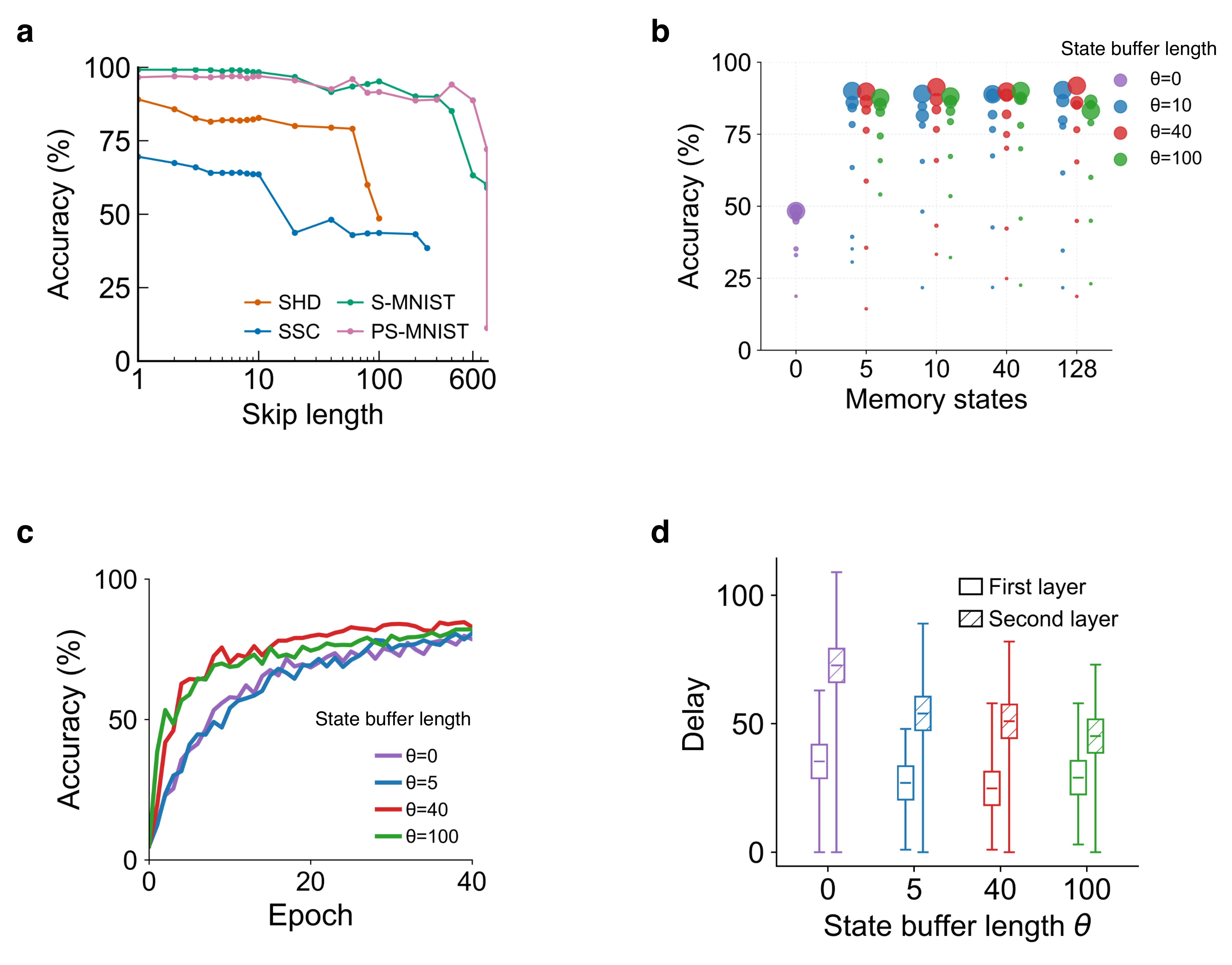}
\caption{\textbf{Context-dependent temporal demands.}
{\textbf{a} Task-dependent slow memory: increasing the memory-update interval (dilation; \emph{x}-axis in log scale) leaves S/PS-MNIST largely unchanged but degrades SHD/SSC, indicating that auditory streams require finer-grained long-term context, whereas long-horizon vision tolerates coarser, less frequent updates.}
\textbf{b} Accuracy versus parameter budget and memory dimension saturates once task-relevant timescales are captured, indicating that capacity should be co-tuned rather than maximised. Bubble size denotes the number of feedforward neurons; colours indicate the state buffer length \(\theta\).
\textbf{c} Increasing the state buffer length ($\theta$) in the DSNN accelerates convergence, consistent with long-range context being carried by the slow memory state.
\textbf{d} Effect of delay distribution with longer state buffer size.  As \( \theta \) grows, probability mass shifts towards shorter delays, the long-delay tail contracts, and overall dispersion decreases-indicating that a longer window partially substitutes for explicit axonal delays.Box plots show the distribution of delays at each state buffer length: the central line represents the mean, the box edges indicate the 25th and 75th percentiles, and the whiskers extend to the minimum and maximum values.}
\label{fig:2_a}
\end{figure}

We next examined how frequently the auxiliary memory must be updated within the dual pathway architecture. To do this, we imposed temporal sparsity by updating the membrane potential over fixed numbers of steps. We found that sequence-vision tasks (S-MNIST, PS-MNIST) were largely unaffected even at a skip length (dilation) of 10, whereas auditory tasks (SHD, SSC) degraded much sooner. Regular dense inputs therefore tolerate coarse updates, while irregular spike streams require tighter coupling  (\textbf{Figure \ref{fig:2_a}a}) which, foreshadowing our later results, is favorable for hardware since less frequent updates lower the switching activity and memory traffic with little loss on long-horizon vision benchmarks. {To make this trend explicit,  \textbf{Extended Data Figure} \ref{maxstate} summarises, for each memory-state size, the {competitive} configurations, showing that strong performance tends to favor relatively short buffers.}

Recent spiking state-space architectures have attempted to remove the feedforward drive (fast pathway) entirely, compensating through residual connections and additional mapping layers ({see \textbf{ Extended Data Figure \ref{sup3_figure}}}). Although this approach is also effective, it substantially increases model complexity - often by up to 3$\times$ in parameters - and imposes additional residual computations that raise hardware cost~\citep{10448152, liulmuformer, stan2024learning,11250792}. In contrast, removing the feedforward pathway in our model caused accuracy to drop to chance across all datasets, irrespective of memory size. { \textbf{Extended Data Table}~\ref{xxx} further reports the accuracy and parameter counts of these model variants, highlighting the improved accuracy-efficiency balance achieved by DMP relative to both the vanilla state-space model and its spiking counterpart.} This indicates that the slow memory state alone cannot sustain task performance. Instead, the explicit temporal state serves as a lightweight contextual modulator that reinforces stimulus-driven spiking rather than acting as a standalone recurrent mechanism. Consistent with this, model capacity on the SHD dataset improved with larger hidden width and memory dimension until task-relevant timescales were captured, beyond which gains saturated (\textbf{Figure~\ref{fig:2_a}b}). These results suggest that neuron count, memory size and temporal horizon should be co-tuned to achieve balanced temporal modelling without excessive computational or hardware overhead.

We next hypothesized that the explicit state already carries long-range context, and therefore should also reduce the need for long axonal delays. Delays are known to align spikes to task-relevant timescales and to improve neuromorphic speech tasks, but long delays require deep buffers that scale with the maximum delay and specialized timing support \citep{hammouamri2023learning,sun2025towards}. Consistent with our hypothesis, we found that combining heterogeneous axonal delays with our memory-augmented neurons improved accuracy for small memory states and, importantly, shifted the learned delays towards shorter horizons. With memory dimension $d=5$ and a maximum delay window $\theta=5$, accuracy increased by 2\%  despite the tight delay budget, and enlarging the delay window mainly accelerated optimization rather than being strictly required (\textbf{Figure \ref{fig:2_a}c-d}). The learned axonal delay distributions are shown in \textbf{ Extended Data Figure \ref{sup1}}.

Together, this points to a hardware-favourable hybrid solution in which the shared temporal state supplies the long-range component without per-connection buffering, while short delays provide fine timing alignment. In this regime, long programmable delays become optional rather than structural, and the network retains the benefits of delay coding, such as robustness to low-precision weights and discretisation \citep{sun2025exploitingheterogeneousdelaysefficient}, and fast motion detection \citep{grimaldi2023learning}, while easing mapping to heterogeneous neuromorphic hardware.

\subsection{Dual memory networks enhance long-term credit assignment}

To assess whether the proposed architecture also facilitates long-range credit assignment, we used last-timestep supervision and quantified how the loss at the final step back-propagates to the first layer over time. In this regime the network must retain task-relevant information across the entire sequence. As shown in \textbf{ Extended Data Figure \ref{sup2}}, gradients in the first layers remained non-negligible over the full 784 steps of S-MNIST and over 200 steps of PS-MNIST, while accuracies stayed high (98.7\% and 96\%, respectively). On SHD the network still propagated useful signal (accuracy about 60\%), whereas SSC proved more challenging, consistent with its more noisy, multiscale temporal structure \cite{alkilany2025neuromodulation}.

A direct comparison with a purely feedforward SNN shows that once the membrane time constant is the only source of memory, gradients vanish rapidly and performance collapses, whereas the slow-memory variant continues to backpropagate through time. This supports the interpretation that adding a slow, local pathway preserves history without having to enlarge the neuron’s intrinsic time constant, and that it can be used as a drop-in component to strengthen existing spiking models that already exploit heterogeneous temporal dynamics.

\begin{figure}[!tbp]
\centering
\includegraphics[width=\linewidth]{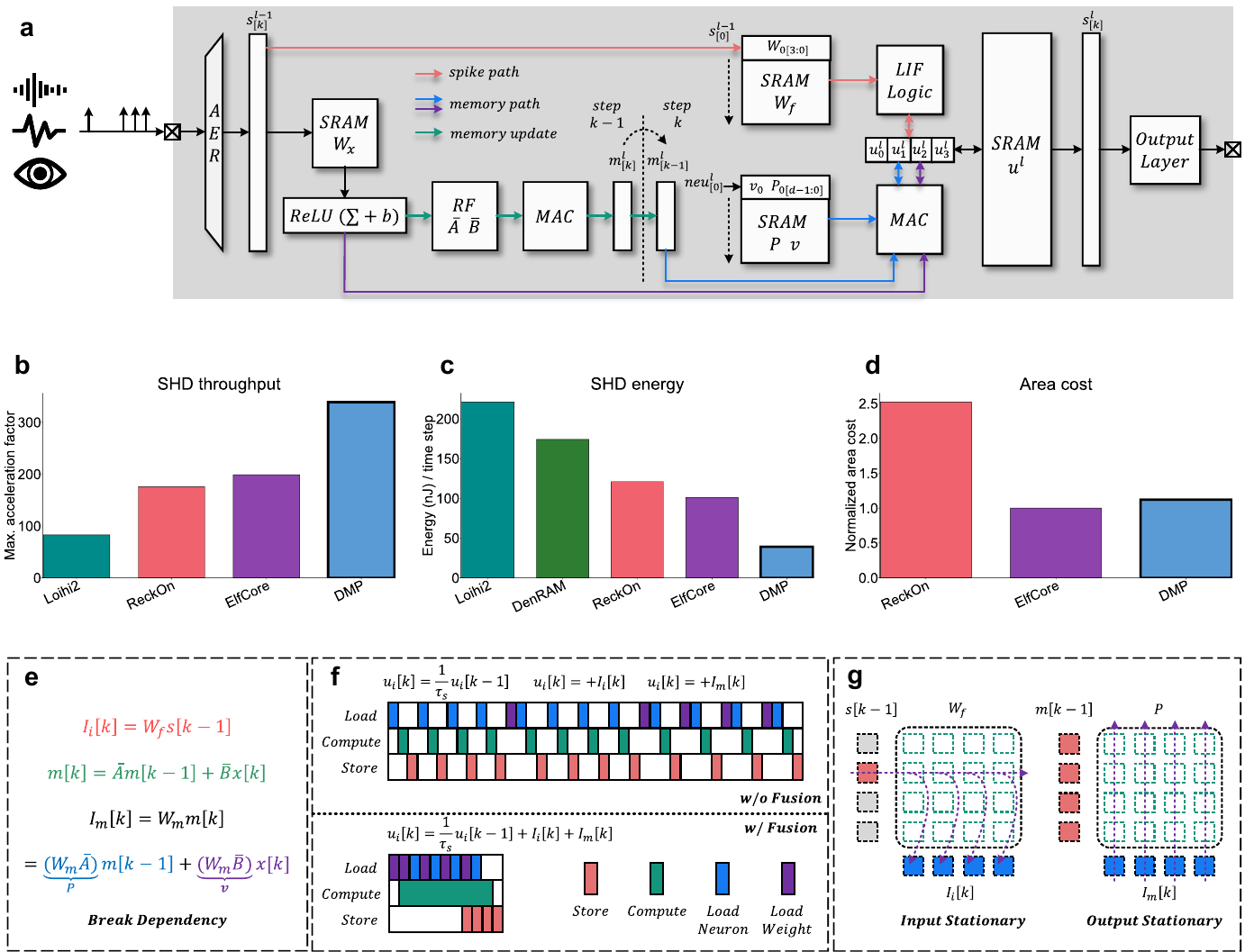}
\caption{\textbf{Hardware design for the dual memory pathway architecture.}
\textbf{a} Hardware architecture for end-to-end temporal, event-based sensory data inference. Four parallel data paths (spike integration, two memory integration paths, and memory update) are shown in distinct colours, enabled by dependency breaking. Dual register slots are allocated for $m[k]$ (memory update) and $m[k-1]$ (memory integration). Four additional register slots temporarily store consecutive neuron membrane potentials $u_l$ for fused LIF operations and vector-matrix multiplications before writing back to neuron SRAM. Weight matrix access patterns differ between spike and memory integration to enhance arithmetic intensity.
\textbf{b} {DMP-SNN} attains the highest throughput per sample by breaking computation dependencies.
\textbf{c} {DMP-SNN} provides up to $5\times$ higher energy efficiency compared with delay-based hardware designs.
\textbf{d} {DMP-SNN} offers $2\times$ greater area efficiency than recurrent SNN architectures, enabled by its reduced parameter-memory footprint.
\textbf{e} Dependency breaking between slow memory update and memory integration enables parallel computation. Precomputed weight memories $P$ and $v$ are stored in SRAM.
\textbf{f} Comparison of data flow with and without fine-grained operator fusion. The example illustrates updates and integrations of four postsynaptic neuron membrane potentials. Operator fusion allows single-access SRAM reads for neuron states, requiring minimal additional register area while significantly increasing arithmetic intensity.
\textbf{g} Heterogeneous operand-stationary schemes are adopted: input-stationary (column-stationary) and output-stationary (row-stationary) access patterns are used for sparse spike integration and dense vector-matrix multiplications, respectively, further improving arithmetic intensity.
}
\label{fig:hardware}
\end{figure}

\subsection{Hardware implementation of dual memory demonstrates superior throughput and energy efficiency}

So far we have shown how the {DMP-SNN} achieves impressive algorithmic performance alongside excellent parameter number scaling on neuromorphic benchmarks. However, these do not yet realize all strengths of {DMP-SNN}, as it was conceived in a co-optimization process to be particularly amenable to hardware implementations. We now introduce a new hardware architecture to realize the full benefits of {DMP-SNN}.

Our hardware architecture is a digital near-memory-compute architecture that allows the co-optimization of dataflow corresponding to the dual memory pathways of the algorithm, instead of the default naive dataflow of existing architectures that would simply integrate the heterogeneous data flows of our algorithm. The end-to-end inference micro-architecture for {DMP-SNN} with a single hidden layer, based on digital near-memory-compute, is depicted in \textbf{Figure~\ref{fig:hardware}a}. On a high level, this architecture supports four parallel computation paths and integrates postsynaptic operations through fine-grained operator fusion and heterogeneous operand stationarity. These design choices break temporal dependencies to balance latency across paths, maximize arithmetic intensity by fusing operators in hardware, and adapt dataflow to handle variable sparsity patterns, collectively minimizing off-chip memory traffic. This modular design scales naturally with task complexity, enabling flexible trade-offs between temporary storage and parallel throughput. The input stage receives task-agnostic binary spike addresses generated by neuromorphic vision or auditory sensors following the address-event representation (AER) protocol~\cite{jetcas_noc}, while the outputs correspond either to classification or regression results, or to spike streams propagated to downstream layers. In the following we first describe the benchmarking results and the following section \ref{sec:hw-optimization} then further details the architectural innovations that made the benchmarking results possible.

We benchmarked DMP-SNN on this co-designed hardware platform against other leading neuromorphic hardware platforms, namely Loihi2~\citep{loihi2}, ReckOn~\citep{reckon}, ElfCore~\citep{elfcore}, and DenRAM~\citep{d2024denram}.
 {The Loihi2 is a programmable digital neuromorphic research platform. The benchmark considered in this paper refers specifically to the delay-based SNN implementation proposed in~\citep{loihi2}.}
{Our architecture achieves over $4\times$ higher throughput than a digital delay-based implementation on Loihi2,}
and over $1.9\times$ higher throughput than recurrent architectures including ReckOn (\textbf{Figure~\ref{fig:hardware}b}). These gains arise from compact slow-memory operations and parallelisation strategies that eliminate data dependencies. 
{Furthermore, DMP-SNN delivers more than $5\times$ higher energy efficiency than both the digital delay-based implementation on Loihi2 and analogue delay-based designs such as DenRAM.}
(\textbf{Figure~\ref{fig:hardware}c}), primarily by eliminating the synaptic-delay buffer overhead (e.g., capacitor arrays in analogue implementations). The integration of operator fusion and heterogeneous operand stationarity yields an additional $2.5\times$ {improvement in energy efficiency over ElfCore, as confirmed in the ablation (see Methods, Experimental setup).} Finally, \textbf{Figure~\ref{fig:hardware}d} shows that  DMP-SNN incurs only a minor area overhead relative to ElfCore due to intermediate accumulation buffers, yet achieves more than $2\times$ higher area efficiency than ReckOn, owing to the removal of recurrent weight-matrix storage. Further details of the hardware validation procedure are provided in \textbf{Methods}. 
{\textbf{ Extended Data Figure~\ref{sup5_figure}} further reports a hardware scaling analysis when doubling the LIF neuron layer.}

Overall, the experimental results demonstrate that DMP-SNN attains high computational efficiency on conventional digital hardware, surpassing architectures fabricated on more advanced technology nodes and even designs leveraging emerging memristive devices.

\subsubsection{Specific hardware optimizations for {DMP-SNN}}
\label{sec:hw-optimization}

Our hardware architecture realizes the full efficiencies of the {DMP-SNN} architecture at the level of advanced data and operator handling optimizations.

\textit{Dataflow-dependency optimization.} Although the compact slow memory state introduces little parameter overhead, its long integration path leads to an imbalanced data flow: the fast spike-integration path can start immediately, whereas the memory-integration path cannot. Therefore, accelerating the memory-integration path is necessary to eliminate the throughput bottleneck. Because the slow memory update is itself a linear operation, the memory integration with the postsynaptic neuron does not need to wait for the update to complete. Instead, the procedure can be divided into two fully parallel stages, as shown in\textbf{ Figure~\ref{fig:hardware}e}. The term on the right, $v\,x[k]$, still depends on the scalar $x[k]$. However, computing the vector-matrix product $P\,m[k-1]$ requires approximately the same amount of time as computing $x[k]$ and then forming $v\,x[k]$. As a result, the computational balance among the critical paths is maintained. In practice, the accumulation of $x[k]$ can be fused with the neuron update $u_i$, ensuring that the scalar $x[k]$ becomes available concurrently with the binary spike vector from the upstream layer.

\textit{Operator-fusion optimization.} According to the {roofline model}~\cite{roofline}, operating in the compute-bound region maximizes arithmetic intensity and hence improves efficiency. In state-of-the-art end-to-end neuromorphic hardware~\cite{reckon, elfcore}, a finite-state machine typically iterates over all neurons to perform leaky updates before spike integration, incurring at least two memory accesses per neuron. The {DMP-SNN} model introduces an additional access for the memory-integration path, further complicating efforts to maximize arithmetic intensity. The hardware implementation fuses leaky updates, spike integration, and memory integration into a single process, as illustrated in \textbf{Figure~\ref{fig:hardware}f}. By using temporary register slots for intermediate accumulation, the design reduces memory access to only one per neuron for the entire sequence of computations, thereby enabling operation in the compute-bound regime.

\textit{Operand-stationarity optimization.} The DMP-SNN algorithm must simultaneously handle spike integration, corresponding to a sparse vector-matrix multiplication (weight accumulation), and memory integration, corresponding to a dense vector-matrix multiplication.  Existing state-of-the-art hardware, such as Kraken~\cite{kraken}, which combines sparse spike-based integration with dense ternary CNN computation, and {ElfCore}~\cite{elfcore}, which integrates sparse spike accumulation with dense eligibility-trace propagation. To optimize both dataflows, {DMP-SNN} adopts heterogeneous operand stationarity~\cite{kwon2019understanding}. For spike integration, we employ {input stationarity} to reduce weight-memory access when the input vector is sparse. For memory integration, we adopt {output stationarity}, minimizing neuron-memory access (only one access per neuron) when the input vector is dense. In this scheme, $W_f$ is indexed by the nonzero entries of the spike vector, whereas $P$ and $v$ are indexed by postsynaptic neuron addresses. As illustrated in \textbf{Figure~\ref{fig:hardware}g}, four consecutive neuron states $u_i$ (or more) are preloaded into register slots for intermediate accumulation. To preserve efficient sparse access for $W_f$, the memory pointers are incremented in a skipped manner based on these four postsynaptic neuron indices.

\section{Discussion}
\label{sec12}
Our work addresses a persistent tension: spiking models require access to long behavioral timescales, yet the mechanisms that typically provide this context - such as dense recurrence or long learnable delays - are precisely those that burden hardware \citep{reckon}. These mechanisms distribute state across many connections, increase memory traffic, and demand deep buffers, causing temporal capability and implementation efficiency to diverge \citep{patino2024hardware}. Here we show that this tension is not inherent and can be resolved through hardware-algorithm co-design. If temporal context is made explicit, compact, and local to each layer, then long-range processing can coexist with hardware efficiency rather than compete with it \citep{roy2019towards}.

On the algorithm level, we draw on a widespread cortical motif in which fast somatic spiking is shaped by slower, local dendritic integration \cite{sartzetaki2025human,cook2025brainlike}. We abstract the biologically observed fast-slow separation into an engineering design principle: a second state that is proximal to the layer but not per-synapse. Each layer maintains a low-dimensional slow state that summarises recent activity and feeds back as an additional current. Being layer-shared, the state’s cost scales linearly rather than quadratically with network width; because its dynamics are well conditioned \citep{voelker2019legendre,gu2020hippo}, it offers a more stable gradient pathway than the decaying membrane. Under strict last-timestep supervision, this configuration sustained useful gradients over hundreds of steps on S-/PS-MNIST, whereas a purely feedforward SNN did not, indicating that long-term credit assignment is carried by the explicit slow state rather than by inflating neuronal time constants.

The hardware mirrors this separation. Instead of forcing sparse spike accumulation and dense linear updates through a single uniform core, we implement two coupled paths: one for spike-driven computation and one for the shared state. Keeping the state on-chip, decoupling update from consume, and adopting distinct dataflow policies for sparse and dense kernels increase arithmetic intensity without extra off-chip traffic. Post-layout results in 22 nm confirm that this mapping holds in silicon: performance matches or exceeds recurrent and delay-based baselines at lower energy and higher throughput, with only a modest area overhead from the additional state. In practice, these hardware gains arise directly from the same design principle used at the algorithmic level: isolate slow information, keep it compact, and move it through a dedicated path. 

{The DMP-SNN hardware targets a different point in the design space than large-scale neuromorphic platforms such as Loihi 2 or SpiNNaker 2. It is primarily designed for edge computing. For large-scale neuromorphic systems, DMP-SNN can serve as an independent core or be integrated into platforms such as Loihi~2 or SpiNNaker. Deploying it on these systems would enable broader evaluation of hardware efficiency and platform compatibility.}
{More importantly, DMP-SNN addresses a key gap in neuromorphic design: the efficient integration of sparse and dense data paths. By incorporating dense data processing into SNNs through heterogeneous hardware integration, it improves model accuracy while maintaining high hardware efficiency.}

Finally, our results suggest a clear algorithm-hardware design principle for long-timescale spiking: make the slow state explicit, shared and low-rank; keep the fast path spike-driven; and execute the two in parallel on hardware. Under this formulation, long temporal horizons no longer compete with efficiency, and the biological fast-slow motif provides a high-level design pattern rather than a loose analogy. This establishes a scalable template for neuromorphic systems in which temporal competence and deployment practicality reinforce each other, rather than exist in trade-off.

\section{Methods}
\label{related}

\subsection*{Spiking neurons}

We model spiking neurons as leaky-integrate-and-fire (LIF) units. For neuron $i$ in layer $l$ at discrete time step $k$, the membrane potential is $u_i^l[k]$. Its dynamics are governed by the membrane decay $\beta$  and the input current $I_i^l[k]$, with the discrete-time update
\begin{equation}
u_i^l[k] \;=\; \beta\, u_i^l[k-1] \;+\; I_i^l[k],
\end{equation}

The input current is
\begin{equation}
I^l[k] = W_f\, s^{l-1}[k]
\end{equation}

where $W_f$ denotes feedforward synaptic weights and $s^{\,l-1}[k]$ the presynaptic spike vector from layer $l{-}1$.

A spike is emitted when the membrane exceeds threshold:
\begin{equation}
s^{l}_{i}[k] = \Theta \big(u^{l}_{i}[k]-\theta_u\big),
\end{equation}
where $\Theta(\cdot)$ is the Heaviside step function (1 if its argument is positive and 0 otherwise), and $\theta_u$ is the membrane threshold.

Building on this vanilla LIF neuron, we consider two variants that extend temporal capacity.
\textbf{LIF with axonal delays} models transmission delays as a temporal shift of presynaptic spikes, i.e., $s^{\,l-1}[k]\rightarrow s^{\,l-1}[k-\bar d]$.
\textbf{LIF with recurrent connections} augments the update by adding an additional recurrent term within the layer.

\subsection*{Slow memory pathway}
We implement the slow pathway using a state-space model that captures long-term dependencies through well-conditioned linear dynamics. {
We treat the slow pathway as a generic {linear memory module} (e.g., linear RNN/SSM-style). In our experiments, we instantiate it with an LMU-like state-space realisation for its explicit parameterisation and stable long-horizon dynamics.} This design builds on the Legendre Memory Unit (LMU) framework \citep{voelker2019legendre,gu2021efficiently}, which formulates a continuous-time memory as a linear time-invariant (LTI) system. The LMU combines a single-input delay network (DN) with a nonlinear dynamical system, efficiently encoding recent input history through orthogonalized basis functions derived from Legendre polynomials.  The delay network projects the input onto a set of orthogonal basis functions derived from Legendre polynomials, providing an efficient representation of temporal context for subsequent nonlinear computation. The DN orthogonalizes the input signal over a sliding window of length $\theta$ (we denote it as state buffer length), while the nonlinear system leverages this orthogonalized memory to compute various functions over time. Mathematically, given an input scalar function $x(t)$, the state update can be described as follows:
\begin{equation}
{m}'(t) = {A}{m}(t) + {B}{x}(t)
\end{equation}
where ${m}(t) \in \mathbb{R}^d$ denotes the memory state vector with dimension $d$, and ${A}$ and ${B}$ are state-space matrices. Following the use of Pad{\'e} approximation \cite{pade1892representation,gu2020hippo}, the state-space matrices can be expressed as follows:

\begin{equation}
\label{pade}
\begin{aligned}
{A} &= \left[ a \right]_{ij} \in \mathbb{R}^{d \times d}, \quad a_{ij} = (2i + 1) \left\{
\begin{array}{ll}
-1 & \text{if } i < j \\
(-1)^{i-j+1} & \text{if } i \geq j
\end{array} \right. \\
{B} &= \left[ b \right]_{i} \in \mathbb{R}^{d \times 1}, \quad b_{i} = (2i + 1)(-1)^{i}, \quad i,j \in [0, d-1]
\end{aligned}
\end{equation}
This continuous-time system can be converted to discrete-time $k$ with a time resolution $\Delta t$:
\begin{equation}
{m}^{l}[k] = {\bar{A}} {m}^{l}[k-1] + {\bar{B}} {x}^{l}[k]
\end{equation}

Here, $\bar{A}$ and $\bar{B}$ are the discretized versions of $A$ and $B$, obtained using the zero-order hold (ZOH) method. Specifically, exact discretization yields
\[
\bar{A} = e^{A \Delta t} \quad \text{and} \quad \bar{B} = A^{-1} (e^{A \Delta t} - I) B,
\]
and these matrices are usually frozen during training.

\subsection*{Dual memory pathway for SNNs (DMP-SNN)}
Each layer maintains a low-dimensional slow state $m^l[k]\in\mathbb{R}^d$ that summarises recent activity and feeds back as an additional current. As suggested by Chilkuri et al. \citep{chilkuri2021parallelizing},  we compress presynaptic spikes into a scalar drive:
\begin{equation}
\label{input}
    {x}^{l}[k] = {f_x(W_{x}s^{l-1}[k]+b)}
\end{equation}

Where $f_x$ is the mapping function. The shared memory is updated by
\begin{equation}
\label{eq:mem-update}
m^l[k] = \bar{A}\, m^l[k-1] + \bar{B}\, x^l[k].
\end{equation}
The  membrane update is
\begin{equation}
\label{eq:mem-u-update}
u^l[k] = \beta\, u^l[k-1] + I^{l}[k] + I_m^l[k]
\end{equation}
 The input current $I$ and memory input $I_m$ are given by:
\begin{equation}
\label{lifI}
I^{l}[k] = W_f s^{l-1}[k],
\end{equation}
\begin{equation}
  I_m^{l}[k] = W_{m}m^{l}[k]
\end{equation}

where $W_m$ represents the weights from memory. This formulation couples a fast spiking pathway with a slow memory that evolves stably over long timescales.  The full detailed equation is listed in \textbf{ Extended Data Table} \ref{tab:lm_dims}. We used the network parameters given in Extended Data Table \ref{tbl:parameter_hyper} for all experiments, unless otherwise specified. Our method is trained via the SpikingJelly and SLAYER frameworks \citep{fang2023spikingjelly, shrestha2018slayer} and utilizes the mean membrane potential of the last layer as the output.

\subsection*{Analysis of the gradient of DMP-SNN}

We analyse temporal credit assignment under last-timestep supervision. For a hidden layer $l$, let $u^l[k] \in \mathbb{R}^N$ and $m^l[k] \in \mathbb{R}^d$ denote the membrane potentials and slow memory at discrete time $k$, respectively. Ignoring input-driven terms (which act as biases and do not affect state-state Jacobians), the dynamics within the layer can be written as
\begin{align}
u^l[k+1] &= \beta\,u^l[k] + W_m\, m^l[k+1], \label{eq:grad_u_update}\\
m^l[k+1] &= \bar{A}\, m^l[k], \label{eq:grad_m_update}
\end{align}
with leak factor $\beta \in (0,1)$, memory readout matrix $W_m \in \mathbb{R}^{N \times d}$, and state-transition matrix $\bar{A} \in \mathbb{R}^{d \times d}$ arising from the state space construction.

It is convenient to group membrane and memory into a joint state
\[
z^l[k] =
\begin{bmatrix}
u^l[k] \\
m^l[k]
\end{bmatrix}
\in \mathbb{R}^{N+d}.
\]
Equations~\eqref{eq:grad_u_update}-\eqref{eq:grad_m_update} define a linear time-invariant update
\[
z^l[k+1] = F\, z^l[k],
\qquad
F =
\begin{bmatrix}
\beta I_N & W_m \bar{A} \\
0         & \bar{A}
\end{bmatrix},
\]
where $I_N$ denotes the $N\times N$ identity matrix. Over a temporal horizon from $k$ to $T$, the same-layer Jacobian is therefore
\begin{equation}
\label{eq:joint_jacobian}
\frac{\partial z^l[T]}{\partial z^l[k]} = F^{\,T-k}.
\end{equation}
Because $F$ is block upper-triangular, its eigenvalues are
\[
\mathrm{spec}(F) = \{\beta\ \text{(multiplicity $N$)}\} \cup \mathrm{spec}(\bar{A}),
\]
where $\mathrm{spec}(\bar{A}) = \{\lambda_i(\bar{A})\}$ denotes the spectrum of $\bar{A}$ and
$\lambda_i(\bar{A})$ is the $i$-th eigenvalue of $\bar{A}$. The eigenvalue $\beta$ arises from the membrane block $\beta I_N$
and the remaining eigenvalues arise from the memory block $\bar A$.
For a purely feedforward SNN (FSNN), the state reduces to the membrane
only, $z^l[k] = u^l[k]$, and $F = \beta I_N$, so that
\[
\frac{\partial u^l[T]}{\partial u^l[k]} = \beta^{\,T-k} I_N,
\]
and all gradient components decay at rate $|\beta|^{\,T-k}$.
In the DMP-SNN, the modes associated with the membrane block still
decay at this fast rate, but the memory block $\bar A$ contributes
additional modes whose decay is governed by $|\lambda_i(\bar{A})|^{\,T-k}$.
Because the slow memory feeds back into the membrane via the term
$W_m m^l[k]$ in Eq.~\eqref{eq:grad_u_update}, these memory modes are
continuously injected into the membrane dynamics and appear as slow
components in $u^l[T]$. By construction, the spectral radius
$\rho(\bar{A})$ is close to one \citep{voelker2019legendre}, so these memory-aligned modes carry
gradients over substantially longer horizons.

Intuitively, the DMP-SNN therefore provides two distinct temporal scales for credit assignment: a fast mode, inherited from the leaky membrane, and a slow mode, inherited from the well-conditioned memory dynamics. Empirically, this leads to a broad temporal tail in the gradient profile over $k$ for the DMP-SNN, whereas the FSNN exhibits a sharp peak near $k=T$ and rapid decay towards earlier time steps (\textbf{ Extended Data Figure \ref{sup2}}).

\subsection*{Dilated slow memory}

Dilated slow memory updates the memory-driven input only every fixed number of time steps, thereby skipping intermediate computations on hardware. The memory-driven input to neuron $n$ at time step $k$ is
\begin{equation}
  I_m^{l}[k] \;=\; W_{m}\, m^{(l)}[k_d],\quad k_d = \lfloor k/d_s \rfloor d_s
\end{equation}
where $d_s$ is the the skip length  (dilation factor), $m^{l}[k_d]$ denotes the memory vector subsampled every $d_s$ steps (i.e., $[\,m[0],\, m[d_s],\, m[2d_s],\,\ldots,\, m(\lfloor k/d_s \rfloor d_s)\,]^\top$). Between updates (when $k \bmod d_s \neq 0$), the membrane potential evolves without the memory injection.

\subsection*{Hardware baselines}
\label{baseline}
SOTA hardware implementations of efficient long-term memory for temporal signal processing include:

{Loihi2}~\cite{loihi2}: An end-to-end framework that enables event-based SNN training with synaptic delays on GPUs and deployment on Intel’s {Loihi2} neuromorphic chip. It represents the SoTA in digital hardware implementations featuring synaptic delays.
{Loihi2 is not specifically optimized for synaptic delay implementation; however, to the best of our knowledge, it provides the only publicly available SoTA results for a digital implementation of SNN delays.}

{DenRAM}~\cite{d2024denram}: An analog feed-forward SNN with dendritic compartments integrated with resistive RAM (RRAM) in 130\,nm technology, supporting both synaptic delays and weights. It represents the SoTA analog hardware approach using memristor-based delayed synapses.

{ReckOn}~\cite{reckon}: A spiking RNN processor that supports task-agnostic online learning over multi-second timescales. It represents the SoTA in digital spiking RNN implementations.

{ElfCore}~\cite{elfcore}: A SNN processor equipped with a local self-supervised learning engine for multilayer temporal learning without labeled inputs. It represents the SoTA in digital hardware integrating both sparse spike and dense trace data paths.
{
For the ablation study, we implemented a modified ElfCore-based architecture without operator fusion or heterogeneous operand stationarity~\cite{elfcore}. 
}

\subsection*{Experimental Setup}
\label{setup}
{

\paragraph{Implementation and Evaluation Methodology}

The proposed DMP-SNN architecture and the ElfCore-based design were implemented in an advanced 22FDX process node and evaluated through post-layout simulations using QuestaSim and Innovus.

Hardware performance metrics were obtained by averaging inference results over the complete SHD test set. All networks were trained offline, and the reported accuracies correspond to hardware-based inference measurements.

\paragraph{Reference Baselines Results}

We compare our design against SoTA neuromorphic platforms, including Loihi2, DenRAM, and ReckOn. The reported metrics for these platforms were extracted from their respective publications and normalized to account for differences in process technology and supply voltage.

\begin{itemize}
    \item \textbf{Loihi2}: Achieves 88\% hardware-measured accuracy and represents the state-of-the-art digital delay-based implementation.
    
    \item \textbf{DenRAM}: Achieves approximately 87\% accuracy and represents the state-of-the-art memristor-based delay implementation. This value is based on hardware-aware simulations rather than full end-to-end silicon measurements.
    
    \item \textbf{ReckOn}: Achieves 86.2\% hardware-measured accuracy and represents a state-of-the-art recurrence-based implementation.

    \item \textbf{ElfCore}: Achieves 90.3\% accuracy on the full SHD test set and serves as the hardware baseline for ablation analysis.
\end{itemize}

The proposed \textbf{DMP-SNN} also achieves 90.3\% accuracy. Both ElfCore and DMP-SNN deploy the identical trained network model. The key difference lies in the architectural enhancements-such as operator fusion and the heterogeneous operator-stationary strategy-which are applied exclusively in DMP-SNN. Therefore, the comparison between ElfCore and DMP-SNN isolates and quantifies the hardware-level benefits introduced by the proposed optimizations.

\paragraph{Comparison Metrics}

The evaluation considers the following hardware-level metrics:

\begin{itemize}
    \item \textbf{Maximum acceleration factor}: The ratio between the core processing speed per time step and real-time execution (1\,ms per SHD time step).
    
    \item \textbf{Energy per time step}: The average energy consumed to process a single inference time step.
    
    \item \textbf{Area cost}: The post-layout, tape-out-ready silicon area of the computing core, reflecting hardware area efficiency under matched (and improved) functionality.
\end{itemize}

\paragraph{Rationale for Selecting ReckOn as the Area Baseline}

ReckOn is chosen as the area comparison baseline for three primary reasons:

\begin{enumerate}
    \item It is the only prior work that reports the complete computing-core area in silicon.
    \item Its architecture closely resembles ours, adopting a single-core design targeted at extreme edge computing scenarios such as sensor-level processing.
    \item It fully utilizes on-chip resources for the same SHD task, ensuring a fair and meaningful area comparison.
\end{enumerate}

}

\section{Data availability}

The spiking data used in this study are publicly available and open source. The dataset for SHD and SSC belong to Spiking Heidelberg Datasets, which can be accessed via \url{https://zenkelab.org/datasets/}. The S-MNIST and PS-MNIST are derived from the original MNIST dataset (\url{http://yann.lecun.com/exdb/mnist/}).


\backmatter

\section{Acknowledgements}
This project was funded by the Advanced Research + Invention Agency (ARIA), was partially supported by the HORIZON EUROPE EIC Pathfinder Grant ELEGANCE (Grant No. 101161114), and has received funding from Swiss National Science Foundation (SNSF 200021E\_222393).
D.A. is funded by an Imperial College Research Fellowship, Schmidt Sciences Fellowship and Templeton World Charity Foundation, Inc (funder DOI 501100011730) under the grant TWCF-2022-30510. J.A. is funded by a Career Development Research Fellowship of St John's College, Oxford.
For the purpose of open access, the authors have applied a Creative Commons Attribution (CC BY) license to the text, figures and code relating to this paper.

\section{Author Contributions}

P.S. and Z.S. contributed equally, designing algorithms, performing experiments, and analysing results. J.A., G.I., D.F.M.G., and D.A. guided methodology, coordinated the study, and all authors contributed to experimental design, analysis, writing, and manuscript review.

\section{Competing interests}
The authors declare no competing interests.

\begin{appendices}

\end{appendices}

\bibliography{sn-bibliography}

\clearpage
\appendix

\section*{ Extended Data Information}
\section*{ Extended Data Figures}
\setcounter{suppfigcounter}{0}
\setcounter{figure}{0}
\renewcommand{\thefigure}{S\arabic{figure}}
\setcounter{table}{0}
\renewcommand{\thetable}{S\arabic{table}}

\suppfigsection{Competitive accuracy across memory-state sizes and buffer lengths}{maxstate}
\begin{figure}[H]
    \centering
    \includegraphics[width=1\linewidth]{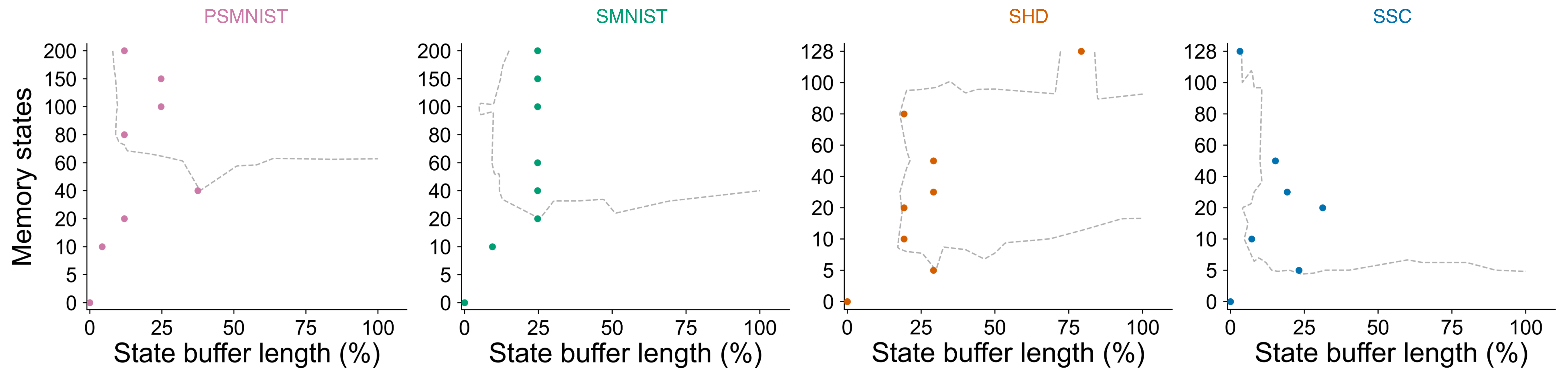}
    \label{dwfce}
\end{figure}

{For each memory-state size, we sweep the state-buffer length and report the {competitive} configurations (within 1.0\% of the best for that size on SHD, and within 0.5\% on the other tasks), for PS-MNIST, S-MNIST, SHD and SSC.
    The grey dashed boundary marks the predefined competitive-performance threshold (defined in the text), highlighting that high performance is typically achieved with compact memory states and relatively short buffers.}
\clearpage

\suppfigsection{Architectural comparison and feedforward ablation}{sup3_figure}
\begin{figure}[H]
  \centering
  \includegraphics[width=\linewidth]{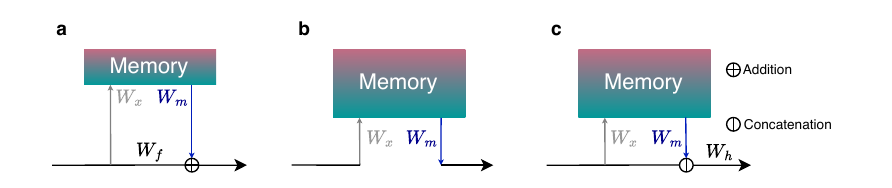}
\end{figure}

{\textbf{a,} Proposed DMP-SNN with a direct feedforward pathway ($W_f$) alongside a shared slow state pathway.
\textbf{b,} Feedforward ablation ($W_f = 0$), yielding a state-only model; performance collapses to chance across tasks, indicating that the slow state acts as a contextual modulator.
\textbf{c,} Representative spiking state-space model (Spiking SSM) that mixes input and state and applies an additional learned transformation ($W_h$), increasing model complexity and compute relative to the DMP design.}

\clearpage

\suppfigsection{Effect of longer state buffer length on axonal delay distributions} {sup1}
\begin{figure}[H]
\centering
  \includegraphics[width=\linewidth]{sup1.drawio_fig3.pdf}
\label{sup1_figure}
\end{figure}

A longer state buffer length ($\theta$) partially absorbs the temporal span that would otherwise be handled by longer axonal delays. The panels show the learned delay distributions as $\theta$ increases while all other settings are held the same. As $\theta$ grows, the probability mass shifts toward shorter delays. The vertical line in the dashed line represents the average delay.

\clearpage
\suppfigsection{Vanilla spiking neurons struggle with long-term credit
assignment} {sup2}
\begin{figure}[H]
  \centering
  \includegraphics[width=1\linewidth]{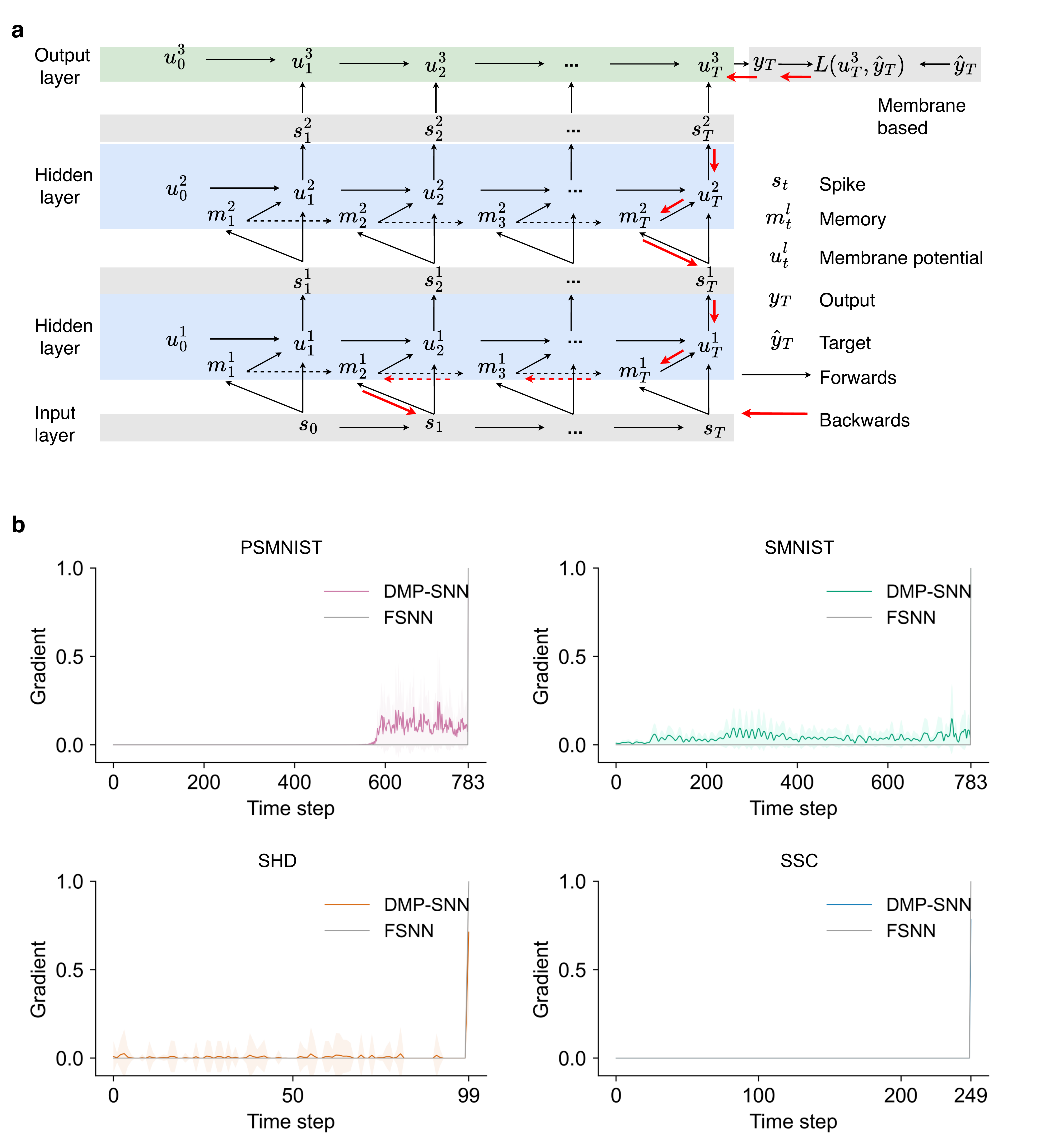}
\label{sup2_figure}
\end{figure}
To clearly assess long-range credit assignment, we adapt the strictly last-timestep supervision. We examined how gradients propagate through time in the dual-memory pathway spiking network (DMP-SNN) and feedforward SNN (FSNN).
\textbf{a,} Computational graph unrolled through time for a spiking neuron during backpropagation through time (BPTT). Red arrows denote the slow gradient pathway visualised in \textbf{b}.
\textbf{b,} Normalised gradient magnitude traced from the terminal loss back to the first hidden layer for PS-MNIST, S-MNIST, SHD, and SSC. Solid lines show the mean and shaded regions indicate the s.e.m. across a random batch (batch size = 256). Under this strict supervision regime, the {DMP-SNN} maintains a broad temporal tail, reflecting stable long-range credit assignment through its slow memory pathway, whereas the feedforward baseline decays rapidly, relying solely on fast membrane dynamics.

\clearpage
\suppfigsection{Hardware scaling analysis} {sup5_figure}
\begin{figure*}[h]
    \centering
    \includegraphics[width=1\linewidth]{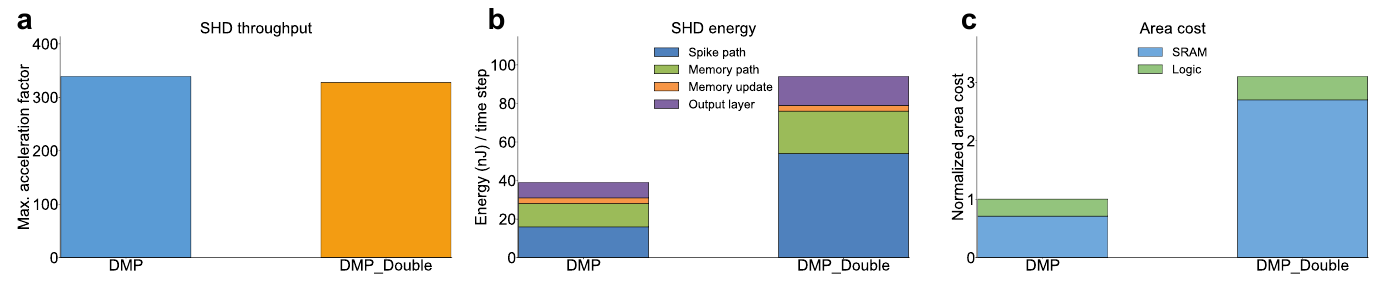}
\label{scaling}
\end{figure*}

{{Hardware scaling analysis when doubling the LIF neuron layer (denoted as DMP\_Double). 
(a) Throughput is largely maintained with $4\times$ MAC and LIF logic, with a slight reduction due to decreased spike sparsity. 
(b) Energy breakdown across spike, memory, memory update, and output paths. Total energy increases by $2.4\times$ (sub-quadratic vs. expected $\sim4\times$), with the spike path becoming dominant ($\approx59\%$). 
(c) Area breakdown into SRAM and logic. Total area increases by $3.1\times$, with SRAM (weights and states) accounting for $\approx87\%$ of the area.}}

{From the algorithmic results in Fig.~\ref{fig:2_a}
(b), we observe that the memory state remains relatively compact, and increasing its size does not necessarily provide additional performance gains. This suggests that, under scaling, variations in the number of LIF neurons have a more direct impact on overall performance compared to scaling the memory state.}

{Accordingly, on the hardware side, we have:
(1) added a detailed cost breakdown of the major components shown in Fig.~\ref{fig:hardware}(a), and
(2) scaled the LIF neuron count to 256 and reported the corresponding hardware metrics.}

{We maintain a single-core design, as extending to a multi-core architecture would require network partitioning, mapping, and routing optimizations, which are beyond the scope of this work. To preserve hardware throughput under scaling, both the MAC array and LIF neuron logic are increased by $4\times$.}

{As shown in  Extended Data Figure ~\ref{sup5_figure}(a-c), we evaluate the scalability of throughput, energy, and area when doubling the size of the LIF neuron layer. The energy breakdown ( Extended Data Figure ~\ref{sup5_figure}(b)) consists of four components corresponding to the data paths in Fig.~\ref{fig:hardware}(a):
\begin{itemize}
    \item \textbf{Spike path:} energy from LIF neuron logic and SRAM access.
    \item \textbf{Memory path:} two parallel paths (blue and purple), accounting for MAC array computation and associated SRAM access.
    \item \textbf{Memory update path.}
    \item \textbf{Output layer.}
\end{itemize}}

{In the area breakdown, SRAM includes storage for both weight parameters and neuron states, while logic includes LIF neuron circuits, MAC units, and ReLU.}

{\textbf{Throughput.}
As shown in  Extended Data Figure~\ref{sup5_figure}(a), the throughput of DMP\_Double is slightly lower than that of DMP, despite the $4\times$ increase in MAC and LIF logic. This is due to slightly reduced spike sparsity in the DMP\_Double configuration.}

{\textbf{Energy.}
As shown in  Extended Data Figure ~\ref{sup5_figure}(b), the total energy of DMP\_Double increases by $2.4\times$ compared to DMP, which is significantly better than the expected $\sim4\times$ increase (assuming quadratic growth in synaptic and MAC operations). The energy scaling factors for the spike path, memory path, memory update, and output layer are $3.4\times$, $1.8\times$, $1\times$, and $1.9\times$, respectively.}

{The spike path becomes the dominant contributor (approximately $59\%$ of total energy in DMP\_Double), which aligns with expectations since only the number of spiking neurons is scaled, while the memory state size remains unchanged. The memory and output paths scale close to $2\times$ due to the increased number of spiking neurons.}

{\textbf{Area.}
The total silicon area of DMP\_Double increases by $3.1\times$ compared to DMP. SRAM-particularly weight storage-gradually dominates the area, accounting for approximately $87\%$ in DMP\_Double. Although the computing logic (MAC and LIF) increases by $4\times$, its area growth remains modest relative to the rapid growth of SRAM area.}

{\textbf{Conclusion.}
The proposed DMP-SNN architecture decouples and parallelizes multiple computation paths, enabling:
\begin{enumerate}
    \item Good throughput scalability (throughput is largely maintained by scaling compute logic).
    \item Sub-quadratic energy growth, even though synaptic and MAC operations increase quadratically.
\end{enumerate}}

{Two key observations emerge:
\begin{itemize}
    \item The area increase of compute logic is negligible compared to SRAM growth.
    \item The spike path increasingly dominates total energy under scaling.
\end{itemize}}

{These findings suggest that further exploration of sparse weight representations in DMP-SNN is a promising direction to reduce both energy and area costs.}

\clearpage

 \section*{ Extended Data Tables}
\captionsetup[table]{name= Extended Data Table, labelsep=colon}
\renewcommand{\thetable}{\arabic{table}}
\setcounter{table}{0}
\captionsetup[table]{name=Table}

\begin{table}[h]
\captionsetup{width=\linewidth}
\centering
\caption{{\textbf{DVS Gesture results.} Test accuracy and parameter count for the baseline feedforward spiking network (FSNN) and the proposed DMP variant.}}
\label{dvsgesture}
\begin{tabular}{lcc}
\toprule
\textbf{Dataset} & \textbf{FSNN} & \textbf{DMP} \\
\midrule
DVS Gesture & 87.12\% / 1.07M & 91.32\% / 1.09M \\
\bottomrule
\end{tabular}
\end{table}

{We evaluated DMP on the DVS Gesture dataset, which comprises event-based spatiotemporal streams and is commonly modelled using spiking convolutional pipelines. As a baseline, we used two spiking convolutional layers followed by two spiking fully connected layers, with pooling applied after each convolution and after the input stage. DMP was inserted into the first spiking fully connected layer, where temporal integration over the extracted features becomes important. All results were obtained with a simulation length of 1000 time steps. As shown in Table~\ref{dvsgesture}, incorporating DMP increases accuracy from 87.12\% to 91.32\% with a marginal increase in parameters (1.07M to 1.09M).}
\clearpage

\begin{table}[h]
\captionsetup{width=\linewidth}
\centering
\caption{{\textbf{Performance comparison across state-space variants.}
Test accuracy (\%, left of each entry) and total parameter count (right; in K) for three architectures: (i) a non-spiking state-space model (SSM), (ii) a spiking state-space model (Spiking SSM), (iii) the proposed dual memory pathway spiking network (DMP-SNN), and DMP removing the feedforward layer($W_f$=0) .
Results are reported on SHD and SSC (event-based auditory classification) and on S-MNIST and PS-MNIST (long-horizon sequential vision).}}

\label{xxx}
\begin{tabular}{lccccc}
\toprule
Dataset & SSM & Spiking SSM &DMP($W_f=0$)  &DMP  \\
\midrule
SHD     & 90.06/73K & 85.09/73K & 5.00/37K & 91.23/40K \\
SSC     & 67.44/75K & 69.13/75K & 2.90/39K& 72.90/42K\\
S-MNIST  & 99.03/202K  & 99.02/202K & 10.00/42K &99.20/73K\\
PS-MNIST  & 97.21/202K & 95.49/202K &10.00/42K&  96.65/102K \\
\bottomrule
\end{tabular}
\end{table}

\clearpage
\begin{table}[t]
\centering
\caption{Dimensions of tensors in the DMP-SNN at layer $l$.
$M$ is the presynaptic width, $N$ the number of postsynaptic neurons, and $d$ the memory dimension.}
\label{tab:lm_dims}
\setlength{\tabcolsep}{6pt}
\renewcommand{\arraystretch}{1.15}
\begin{tabular}{l l l}
\toprule
\textbf{Symbol} & \textbf{Role} & \textbf{Dimension} \\
\midrule
$s^{\,l-1}[k]$  & Presynaptic spikes                     & $\{0,1\}^{M}$ \\
$u^{\,l}[k]$    & Membrane potentials (postsynaptic)     & $\mathbb{R}^{N}$ \\
$m^{\,l}[k]$    & Slow memory state                    & $\mathbb{R}^{d}$ \\
$x^{\,l}[k]$    & Memory input                            & $\mathbb{R}^{1}$ \\
$I^{\,l}[k]$    & Input current                           & $\mathbb{R}^{N}$ \\
$I^{\,l}_{m}[k]$    & Memory current                      & $\mathbb{R}^{N}$ \\
\midrule
$W_f$     & Feedforward ($s\!\rightarrow\!u$)      & $\mathbb{R}^{\,N \times M}$ \\
$W_m$     & Memory readout ($m\!\rightarrow\!u$)   & $\mathbb{R}^{\,N \times d}$ \\
$\bar A$  & Memory transition (state matrix)       & $\mathbb{R}^{\,d \times d}$ \\
$\bar B$  & Memory input map ($x\!\rightarrow\!m$) & $\mathbb{R}^{\,d \times 1}$ \\
$W_x$     & Spike compression ($s\!\rightarrow\!x$)& $\mathbb{R}^{\,1 \times M}$ \\
$b$       & Bias for $x$                           & $\mathbb{R}^{1}$ \\
\midrule
$W_r$ & Recurrent ($s\!\rightarrow\!u$) & $\mathbb{R}^{\,N \times N}$ \\
$\bar d^{\,l}$ & Axonal delays (per output channel)        & $\mathbb{R}^{\,N }$\\
\bottomrule
\end{tabular}
\end{table}

\clearpage

\begin{table*}[t]
    \centering
    \caption{Hyperparameter settings for different datasets.}
    \label{tbl:parameter_hyper}
    \begin{tabular}{lcccc}
        \hline
        \textbf{Parameter}
          & \textbf{SHD}
          & \textbf{SSC}
          & \textbf{S/PS-MNIST}
          & \textbf{Description} \\
        \hline
        Input channels
          & 140
          & 140
          & 1
          & Number of input units  \\
        Simulation Time Steps
          & 100
          & 250
          & 784
          & Simulation Time Steps \\
        Output neurons
          & 20
          & 35
          & 10
          & Number of classes \\
        Hidden Spiking Neurons
          & 128
          & 128
          & 200
          & Size of hidden layer  \\
        Memory states
          & 10
          & 10
          & 40
          & Dimension of Memory states  \\
        State buffer length
          & 40
          & 40
          & 300
          & Sliding window of length ($\theta$)  \\
        Number of Hidden Layers
          & 2
          & 1 and 2
          & 2
          & Number of hidden layers  \\
\hline
        $\theta_u$
          &
          & 10
          &
          & Spiking threshold \\

      $f_x$
          &
          & ReLU
          &
          & Non-linear function \\

        $\bar d$
          &
          & learnable
          &
          & Axonal delays \\
        \hline
    \end{tabular}
\end{table*}

\end{document}